\documentclass[english, letterpaper, 10pt]{article}

\usepackage[T1]{fontenc}
\usepackage{textcomp}

\usepackage{graphicx}
\usepackage{amssymb,amsmath}
\usepackage{fixltx2e}
\usepackage{fancyhdr}
\usepackage{url}
\usepackage{makecell}
\usepackage{subfigure}
\usepackage{caption}

\usepackage[numbers,sort&compress]{natbib}

\usepackage{dsfont}
\usepackage[top=1in, bottom=1.25in, left=0.875in, right=0.875in]{geometry}
\usepackage{bm}

\DeclareMathOperator*{\argmin}{arg\,min}
\DeclareMathOperator*{\argmax}{arg\,max}

\usepackage{mathptmx}   % turns on Times New Roman font
\usepackage{titlesec}
\usepackage{titling}

\usepackage{booktabs}
\usepackage{graphicx}
\usepackage{multirow}

\usepackage[english]{babel}
\usepackage[utf8]{inputenc}
\usepackage{amsmath}
\usepackage{amsfonts}
\usepackage{graphicx}
\usepackage{enumitem}
\usepackage[colorinlistoftodos]{todonotes}
\usepackage{abstract}
\newcommand{\txt}[1]{\textnormal{#1}}

\usepackage{indentfirst}%Indent in first paragraph

\usepackage{times}%switch to Times New Roman font

\usepackage{titlesec}% formatting of section title
\titleformat{\section}{\bfseries\centering\fontsize{11pt}{13pt}\selectfont}{\thesection.}{4pt}{\uppercase}
\titleformat{\subsection}{\bfseries\fontsize{11pt}{13pt}\selectfont}{\thesubsection}{6pt}{}
\titlespacing*{\section}{0pt}{10pt plus 8pt}{4pt}
\titlespacing*{\subsection}{0pt}{6pt}{3pt}

\usepackage{titling}% formatting of article title
\pretitle{\vspace{-15mm}\begin{center}\bfseries\LARGE}
\posttitle{\end{center}}
\preauthor{\begin{center}\large}
\postauthor{\end{center}\vspace{-15mm}}

\usepackage{etoolbox}
\apptocmd{\thebibliography}{\setlength{\itemsep}{-2pt}}{}{}

\newlength{\myitemsep}
\setlist[itemize]{itemsep=-1pt, topsep=2pt}
\setlist[enumerate]{itemsep=-1pt, topsep=1pt}

\title{Meta-Learning\\for Resampling Recommendation Systems\thanks{%The research, presented in Section \ref{sec:exp-res-recsyst} of this paper, was partially supported by the Russian Foundation for Basic Research grants 16-01-00576 A and 16-29-09649 ofi m. 
The research was supported solely by the Ministry of Education and Science of Russian Federation, grant No. 14.606.21.0004, grant code: RFMEFI60617X0004.}
}

\author{Dmitry Smolyakov$^{1}$, Alexander Korotin$^{1}$, Pavel Erofeev$^{2}$, Artem Papanov$^{2}$, Evgeny Burnaev$^{1}$
\\~\\
$^{1}$Skolkovo Institute of Science and Technology\\
\textit{Nobel street, 3, Moscow, Moskovskaya oblast’, Russia}
\\~\\
$^{2}$Institute for Information Transmission Problems\\
\textit{Bolshoy Karetny per. 19, build.1, Moscow, Russia}}

\date{}

\begin{document}
\pagenumbering{gobble}
\maketitle

\renewcommand{\abstractname}{\vspace{0pt}\fontsize{11pt}{13pt}\selectfont \uppercase{Abstract}\vspace{-4pt}}
\begin{abstract}
\normalsize
One possible approach to tackle the class imbalance in classification tasks is to resample a training dataset, i.e., to drop some of its elements or to synthesize new ones. There exist several widely-used resampling methods. Recent research showed that the choice of resampling method significantly affects the quality of classification, which raises the resampling selection problem. Exhaustive search for optimal resampling is time-consuming and hence it is of limited use. In this paper, we describe an alternative approach to the resampling selection. We follow the meta-learning concept to build \emph{resampling recommendation systems}, i.e., algorithms recommending resampling for datasets on the basis of their properties.
\vspace{0.1mm}
\end{abstract}
\section{Introduction}
\label{sec:intro}
In this paper, we consider the task of classification on imbalanced datasets. 
This is a special case of two-class classification task when one class, `minor~class', has much less representatives in the available dataset than the other class, `major~class'.
These conditions are of a big interest because many real-world data analysis problems have inherent peculiarities which lead to unavoidable imbalances in available datasets.
Examples of such problems include network intrusion detection (\citet{net-intrusion}, \citet{smol}), oil spill detection from satellite images (\citet{oilspill}, \citet{ignat}), detection of fraudulent transactions on credit cards (\citet{cardfraud}), diagnosis of rare diseases (\citet{med}), prediction and localization of failures in technical systems (\citet{faultloc,pm}), etc.
These and many other examples have one common significant feature: target events (diseases, failures, etc.) are rare and therefore they generally constitute only a small fraction of available data.
Hence an attempt to naturally formulate any of these problems as a binary classification task
(target events form one class, no-events form another class) leads to the class imbalance.
Note that this effect is unavoidable since it is caused by the nature of the problem.

Moreover, according to \citet{learn-imbalanced}, the accurate detection of minor class elements is often the main goal in imbalanced problems.
In the above-mentioned examples, minor class corresponds to target events whose accurate prediction is crucial for applications.
However, standard classification models (Logistic regression, SVM, Decision tree, Nearest neighbors) treat all classes as equally important and thus tend to be biased towards major class in imbalanced problems (see \citet{svm-unbalanced},  \citet{logreg-unbalanced},  \citet{dtree-unbalanced-1},  \citet{dtree-unbalanced-2}).
This may lead to inaccurate detection of minor class elements with the average quality of prediction being high.
E.g., consider a process with events occurring just $1$\% of all times.
If a classification model always gives a `no-event' answer it is wrong in just $1$\% of all cases. The average quality of the classifier is good.
But such prediction is useless for minor class detection.
Thus, imbalanced classification problems require special treatment.
% in terms of quality assessment and classifier learning
% special quality measures

One possible way (\citet{learn-imbalanced}) to increase importance of minor class and deal with peculiarities described above is to \emph{resample} dataset in order to soften or remove class imbalance.
Resampling may include:
\emph{oversampling}, i.e., addition of synthesized elements to minor class;
\emph{undersampling}, i.e., deletion of particular elements from major class;
or both.
Resampling is convenient and widely used since it allows to tackle imbalanced tasks using standard classification techniques.
On the other hand, it requires the selection of the resampling method.

There exist several \emph{resampling methods}, i.e. algorithms describing which observations to delete and how to generate new ones.
Most of them take the \emph{resampling amount} as an argument, which governs how many observations are added or deleted.
Thus, to apply resampling to a classification task, one has to select a method and a resampling amount.
%and, for some resapling methods, also additional parameters

Previous research by \citet{res-influence} showed that there is no `universal' choice of resampling method and resampling amount that would improve classification quality for all imbalanced classification tasks.
%! Add: This is consistent with No-Free-Lunch theorem
Moreover, the choice of resampling method strongly affects the quality of the final classification. This influence varies from one dataset to another.
Therefore, one has to select resampling method and resampling amount properly for every particular task.

The straightforward approach to select the optimal resampling method and amount is to perform the exhaustive search. One tries all considered options, estimates their quality using cross-validation on the train data, and selects the method with the highest quality.
It is ideologically simple and reasonable, but it is also time-consuming.
Thus exhaustive search is of limited use.

In this work, we explore an alternative approach to resampling selection problem.
We develop \emph{resampling recommendation systems}, that are algorithms which recommend resampling for datasets on the basis of their properties (statistical, metric, etc.).
We aim to construct recommendation systems that satisfy two key requirements.
Firstly, they have to be significantly faster than cross-validation exhaustive search.
Secondly, they have to give better quality (see section~\ref{sec:exp-res-recsyst-accuracy} for details) than trivial resampling selection strategies, such as using the same resampling method for each dataset.
%! reference to section in the middle of the paper --- should it be here?
%Note that we do not aim to build an ideal rec.system

To construct resampling recommendation systems, we follow the \textit{meta-learning} approach.
We analyze experience of applying resampling methods to various imbalanced classification tasks and used it to \emph{learn} recommendation systems.
Namely, we take a big number of various classification tasks on imbalanced datasets and calculate their characteristics.
We apply various resampling methods with various resampling amount values to every task, learn classification models on resampled datasets and estimate their quality.
Each task is a learning example (\emph{meta-example}), its characteristics act as its features (\emph{meta-features}), quality values act as the target variables.
We use the set of meta-examples as the training dataset to learn resampling recommendation systems (for details see section~\ref{sec:res-recsyst-construct}).

In this paper, we describe construction and evaluation of resampling recommendation systems.
\textbf{The main contributions of the article are:}
\begin{enumerate}
\item
Developing two natural approaches to build resampling recommendation systems using meta-learning.

\item
Learning recommendation systems on numerous artificial and real-world imbalanced classification tasks.\footnote{Full source code and the datasets (used in the experiment) are available online in Public GitHub repository \url{https://github.com/papart/res-recsyst}.}

\item
Exploring the performance of the recommendations systems. We show that they give satisfactory results while demanding much less time than exhaustive search.
\end{enumerate}

In section~\ref{sec:imb-clf-task} we formulate the task of binary classification on imbalanced datasets and give a brief overview of resampling methods.
Section~\ref{sec:resamplers-compare} summarizes results of experimental resampling methods comparison by \citet{res-influence}.
%and thus gives motivation for further discussion
Section~\ref{sec:recsyst} provides an overview of previous works on recommendation systems.
In section~\ref{sec:res-recsyst-construct}, we describe two approaches to construct resampling recommendation systems.
Finally, in section~\ref{sec:exp-res-recsyst} we provide all the details of the experimental part of our research, evaluate the quality of resampling recommendation systems and discuss the results.

\section{Imbalanced Classification Task}
\label{sec:imb-clf-task}

\subsection{Notation and Problem Statement}
\label{sec:imb-notation}
Consider a dataset with $\ell$~elements $S = (X_i, y_i)_{i = 1}^{\ell}$, where $X_i \in \mathbb{R}^d$ is an instance of $d$-dimensional feature space and $y_i \in \{0, 1\}$ is a class label associated with $X_i$.
Denote $C_0(S) = \{(X_i,y_i)\in S \mid y_i=0\}$ and $C_1(S) = \{(X_i,y_i)\in S \mid y_i = 1\}$.
Let label $0$ correspond to the major class, label $1$ correspond to the minor class, then $|C_0(S)| > |C_1(S)|$.
To measure a degree of class imbalance for a dataset, we introduce an \emph{imbalance ratio}
$I\!R(S) = \frac{|C_0(S)|}{|C_1(S)|}$.
Note that $I\!R(S) \geq 1$ and the higher it is, the stronger imbalance of $S$ is.

The goal is to learn a classifier using training dataset~$S$.
This is done in two steps.
Firstly, dataset $S$ is \emph{resampled} by \emph{resampling method} $r$ with \emph{resampling multiplier} $m>1$: some observations in $S$ are dropped or some new synthetic observations are added to~$S$.
Resampling method $r$ determines how observations are deleted or how new ones are synthesized.
Multiplier $m$ governs resampling amount by setting the resulting imbalance ratio as $\frac{1}{m} I\!R(S)$.
Thereby the result of resampling is a dataset
$r_m(S)$ with $I\!R(r_m(S)) = \frac{1}{m}I\!R(S) < I\!R(S)$.
Secondly, some standard classification model $h$ is learned on $r_m(S)$, which gives classifier
$h_{r_m(S)}: \mathbb{R}^d \rightarrow~\{0, 1\}$ as a result.

Performance of a classification model with resampling $(r_m, h)$ is determined by a predefined \emph{classifier quality metrics} $Q(h_{r_m(S_{\txt{train}})},S_{\txt{test}})$.
It takes as input classifier $h_{r_m(S_{\txt{train}})}$ learned on resampled training dataset, testing dataset~$S_{\txt{test}}$ and yields higher value for better classification.
In order to determine performance of $r_m$ and $h$ on the whole dataset~$S$ regarding metrics~$Q$, we use a standard procedure based on $k$-fold cross-validation \citet{stat-learning} which yields value of $Q$ for each CV-iteration: $Q^{kCV}(h, r_m, S) \in \mathbb{R}^k$.
We will consider their arithmetic mean $Q^{kCV}_{\txt{av}}(h, r_m, S) = \overline{Q^{kCV}(h, r_m, S)}$ as a quality estimate, but single components of $Q^{kCV}$ will be also used.

It is convenient to regard identity transformation of a dataset as a trivial resampling method.
We call it ``no-resampling'' and denote it as $r^0$, then, by definition, $r^0_m(S) = S$ for any multiplier $m$.

\subsection{Resampling Methods Overview}
\label{sec:resamplers}

Every resampling method $r$ considered in this paper works according to the following scheme.

\begin{enumerate}
\item
Takes input: dataset $S$ (as described in section \ref{sec:imb-notation}), \emph{resampling multiplier} $m>1$ which determines resulting imbalance ratio as $I\!R(r_m(S)) = \frac{1}{m} \cdot I\!R(S)$ and thereby controls resampling amount, additional parameters (specific for every particular method).

\item
Modifies given dataset by adding synthesized objects to the minor class (\emph{oversampling}), or by dropping objects from the major class (\emph{undersampling}), or both. Details depend on the method used.

\item
Outputs
resampled dataset $r_m(S)$ with $d$ features and new imbalance ratio ${I\!R(r_m(S)) = \frac{1}{m} \cdot I\!R(S)}$.

\end{enumerate}

In this paper, we consider three most widely used resampling methods: Random Oversampling, Random Undersampling and Synthetic Minority Oversampling Technique (SMOTE).

\subsubsection{Random Oversampling}

Random oversampling (ROS, also known as bootstrap oversampling, see \citet{learn-imbalanced}) takes no additional input parameters.
It adds to the minor class new $(m-1)|C_1(S)|$ objects.
Each of them is drawn from uniform distribution on $C_1(S)$.

\subsubsection{Random Undersampling}
Random Undersampling (RUS, see \citet{learn-imbalanced}) takes no additional parameters.
It chooses random subset of $C_0(S)$ with $\frac{m-1}{m}\cdot |C_0(S)|$ elements and drops them from the dataset.
All subsets of $C_0(S)$ have equal probabilities to be chosen.

\subsubsection{SMOTE}

Synthetic Minority Oversampling Technique (SMOTE, see \citet{smote}) takes one additional integer parameter $k$ (number of neighbors).
It adds to the minor class new synthesized objects, which are constructed in the following way.
\begin{enumerate}
\item
Initialize set as empty: $S_{\txt{new}} := \emptyset$

\item
Repeat the following steps $(m-1)|C_1(S)|$ times:
\begin{enumerate}[label=(\roman*)]
\item
Randomly select one element $X_i \in C_1(S)$.

\item
Find $k$ minor class elements which are nearest neighbors of $X_i$.
Randomly select one of them and denote it by $X_j$.

\item
Select random point $x$ on the segment [$X_i, X_j]$.

\item
Assign minor class label to the newly generated element $x$ and store it: $S_{\txt{new}} := S_{\txt{new}} \cup \{(x, 1)\}.$
\end{enumerate}

\item
Add generated objects to the dataset: $\tilde{S} = S\cup S_{\txt{new}}$.
\end{enumerate}

\subsubsection{Other Resampling Methods}

There exist several other resampling methods:
Tomek Link deletion by \citet{one-sided-selection}, One-Sided Selection by \citet{one-sided-selection}, Evolutionary Undersampling by \citet{eus}, borderline-SMOTE by \citet{borderline-smote}, Neighborhood Cleaning Rule by \citet{ncl}.
There exist also procedures combining resampling and classification in boosting:
SMOTEBoost by \citet{smoteboost}, RUSBoost by \citet{rusboost}, EUSBoost by \citet{eusboost}.
%!Add: bagging methods
These methods are not examined in this paper and not considered in further resampling recommendation system construction,
but methodology we describe is quite general, so it can be applied to these methods as well.

\section{Influence of Resampling on Classification Accuracy}
\label{sec:resamplers-compare}

The influence of resampling on the classification accuracy was explored experimentally by \citet{res-influence}.
The authors measured accuracy of various classification models on several artificial and real-world imbalanced datasets, which were preliminarily resampled using one of three methods (Random oversampling, Random undersampling and SMOTE) with multiplier $m$ from 1.25 to 10.
Analyzing results of these experiments, the authors came to the following conclusions.

Resampling can have both positive and negative effects on the classification quality.
The effect strongly depends on the selected resampling method and multiplier.
If the method and the multiplier are selected properly, resampling can significantly improve classification accuracy in most cases.
Still, in some cases, classification without resampling is the best choice.
In addition, impact of resampling on quality depends on the data it is applied to, so there is no method that would guarantee quality improvement for all datasets.

\section{Recommendation Systems Related Work}
\label{sec:recsyst}

The results by \citet{res-influence} raise the problem of resampling method and multiplier selection.
In fact, proper selection of the method and its parameters is crucial for many other types of data analysis tasks.
One possible approach to this problem is to use meta-learning to build systems aimed to recommend method and parameters for task solving.
In this section, we discuss construction, application and evaluation of meta-learning-based recommendation systems.

\subsection{Recommendation System: Motivation}
\label{sec:rec-motivation}
%%SHORT: suppose proper method selection is needed; exhaustive search is heavy; alternative - rec. syst.
It is a common case when certain type of data analysis tasks (e.g., classification, time series prediction, feature selection) has plenty of methods for its solution.
Suppose that some quality metric is given, so solutions obtained by different methods can be compared.
The choice of a method may strongly affect the quality of the solution (e.g., for resampling methods it is demonstrated by \citet{res-influence}).
In that case, it becomes necessary to select the method properly.
The straightforward way to do this is to perform exhaustive search.
However, this approach is time-consuming, so its use is practically impossible in some cases.
Trying to overcome this problem, many researchers explored an alternative approach to method selection recommendation systems.

%%SHORT: def of rec.syst.; 2 requirements of rec.syst.; researcher sets the trade-off
By a \emph{recommendation system} for a certain type of data analysis tasks we mean an algorithm which takes as input a task (its description and its training dataset) and gives as output a \emph{recommendation} on how to solve this task.
%!ref to examples of such systems
The recommendation may be detailed in different ways: it may contain only one method which is considered by the system as the best, or provide ranking of methods preferable for the task, or also provide values of parameters for the recommended methods, etc.
%rewrite
Recommendation systems aim to fulfill two requirements. They have to provide recommendations of high quality and they have to be computationally fast.
These requirements are conflicting, so a researcher has to decide which one is of higher priority or what trade-off between them is preferable.
%E.g., in this paper, we give priority to the latter requirement and then try to achieve better recommendation quality.
%!why? because for most accurate decision there is exhaustive search
%That is, we aim to build a recommendation system which will take much less time than exhaustive search to provide reasonable (but not necessarily optimal) recommendations.
%There are other approaches, see sec. ...

%%SHORT: bridge: concept -> constuction & quality estimation
%This concept raises two questions:
%how to construct recommendation systems and how to estimate their quality.

%%SHORT: meta-learning for rec.syst. construction; refs to subsections
One possible approach to recommendation system construction is \emph{meta-learning}.
It has been successfully used for various types of tasks (see section \ref{sec:rec-examples} for examples)
%!ref to sec
and in the paper we also follow this approach.
In subsection~\ref{sec:meta-learning} we describe meta-learning concept and in subsection~\ref{sec:rec-template} we give the template for meta-learning-based recommendation system construction.

\subsection{Meta-Learning Concept}
\label{sec:meta-learning}
%%SHORT: meta-learning idea; def meta-features, -examples, etc.; learn algorithm for quality-variables prediction
\emph{Meta-learning} is a widely-used approach to recommendation system construction.
According to \citet{rec-ml-concepts}, its idea is that there is some intrinsic relation between task properties and performance of methods for its solution,
so one can try to extract this relation and use it to build a recommendation system.
Generally, the extraction of this relation is a challenging problem. There exists many possible approaches.
%!ref?
In the case of meta-learning tasks of the same type are considered as learning examples and are called \emph{meta-examples}, their properties act as features and are called \emph{meta-features}, quality values of methods applied to each meta-example (\emph{quality-variables}) are its target variables (\emph{target quality variables}).
%!called, called, called --- repetitions!
That is, one takes a set of tasks of the same type, calculates their meta-features, applies methods to each of them, computes their quality estimates, gathers them in quality-variables and thereby forms a \emph{training meta-dataset}.
%!heavy sentence
This dataset is used to learn a model predicting quality-variables by meta-features and hence representing the required relation.

%%SHORT: decision function: quality-variables -> recommendation
When the model is learned, it can predict quality-variables for any task, either from the training meta-dataset or an unseen one.
After that, we choose a decision function which will convert these predicted values into a recommendation.

%%SHORT: requirements to rec.syst. -> requirements to algorithm accuracy; meta-features, algorithm, dec.func. complexity
The requirements to recommendation system  (see section~\ref{sec:rec-motivation}) in meta-learning terms mean the following.
Firstly, the model predicting quality-variables using meta-features should be as accurate as possible.
Secondly, meta-features calculation, quality-variables prediction, decision function application have to be computationally cheap.
%Recall that in our paper we start from the second requirement and then try to fulfill the first.

%%SHORT: refs to template and details of realization
Note that the meta-learning concept does not presume specific realization.
In this paper, we follow a template described in the following subsection~\ref{sec:rec-template}.
Further details of resampling recommendation systems we build are described in section~\ref{sec:res-recsyst-construct}.

\subsection{Recommendation System Template}
\label{sec:rec-template}

%%SHORT: steps of the template, notation
In this subsection, we describe a template for meta-learning-based recommendation system realization.
It is quite general and covers various recommendation systems from previous works (for details see section~\ref{sec:rec-examples}). Resampling recommendation systems we built also fit this template.
The template is given below.

\begin{enumerate}
\item
\emph{Decide the type of considered tasks.}
%%%We consider imbalanced binary classification tasks.

\item
\emph{Decide the sets of considered methods and parameters.}
%%%According to section~\ref{sec:imb-notation}, to solve a classification task, one selects resampling method $r$, multiplier $m$ and classification model $h$.
%%%In further consideration, classification model $h$ is fixed
%%%and we aim to select resampling method and multiplier from sets $R=\{r^0,\ r^1, \ldots, r^{n_1}\}$ and $M=\{m_1, \ldots, m_{n_2}\}$, respectively.

\item
\emph{Decide the form of recommendation.} 
E.g., single method, several methods, or ranking of methods.
%%%We aim to recommend one pair $(r\in R, \ m\in M)$.

\item
	\emph{Select a quality metric} which will be used to estimate performance of each method on every task.
%%%We fix classification quality metric $Q$, number of cross-validation folds $k$.
%%%Components of $Q^{kCV}(h, r_m, S)$ will compose quality-variables,
%%%and their average $Q^{kCV}_{av}(h, r_m, S)$ will be used as an overall quality estimate on dataset $S$.

\item
\emph{Select a recommendation accuracy metric} which will be used to assess the recommendation system on each task.
It can be composed from quality of methods recommended by the system, quality of other methods, runtime of methods.
Further discussion of this metric is provided in the following subsection~\ref{sec:rec-accuracy}.

\item
\emph{Formulate the meta-learning problem.}
%It is a key point in construction of a recommendation system because it determines which exactly solving experience will be used in meta-learning and what exactly for.

	\begin{enumerate}%[label=(\roman*)]
    \item
    \emph{Define meta-features.}
    We denote meta-features of a task with dataset $S$ as $f(S)$ and number of them as $n_f$.
    %rewrite
    
    \item
    \emph{Define quality-variables} which aggregate information about performance of methods on the meta-example.
    We denote a vector of quality-variables of the task with dataset $S$ by $q(S)$ and a number of them by $n_q$.

    %%%For any dataset $S$, $k$-dimensional vector $Q^{kCV}(h, r_m, S)$ will be calculated for each $r \in R$, each $m \in M$, which gives $k n_m (n_1 + 1)$ values in total.
    %%%Thus we define a function $\tilde{q} \colon \mathbb{R}^{k n_m (n_1 + 1)} \rightarrow \mathbb{R}^{n_q}$ where $n_q$ is a number of quality-variables.
    %%%We denote vector of quality-variables of dataset $S$ as $q(S)$, i.e.
    %%%$q(S) = \tilde{q}\left( \left.Q^{kCV}_{av}(h, r_m, S)\right|_{r \in R, \ m \in M} \right)$. 
    \item
    \emph{Define target quality-variables}, i.e. some of quality-variables which are to be predicted using meta-features.
	Note that not every quality-variable is supposed to be a target in the meta-learning problem.
	Some of them can serve as auxiliary variables used to form the targets or to estimate quality of the recommendation.
    \end{enumerate}

\item
\emph{Construct the training set of meta-examples.}
	\begin{enumerate}%[label=(\roman*)]
	\item
	Collect different tasks of the same type. They will compose a training meta-dataset.
  Denote this bank of tasks by 
  $B=\{S_1,\ldots,S_{n_S}\}.$

	\item
  Calculate the meta-features of every training meta-example: 
  $\{f(S_1), \ldots, f(S_{n_S})\}.$

	\item
  Apply each considered method to every task and get values of quality measure.

    %%%For our problem, we calculate $Q^{kCV}(h, r_m, S)$ for each $r \in R$, $m \in M$, $S \in B$.

  \item
  Calculate the values of quality-variables for each meta-example in the set: 
  $\{q(S_1), \ldots, q(S_{n_S})\}.$

  %%%For each dataset $S \in B$ we get $n_q$ quality-variables $q(S)$ as it has been described above.
  \end{enumerate}

\item
\emph{Solve the meta-learning problem} by fitting some model on the set of training meta-examples.
This model is represented as function $p(f(S))$ taking values of meta-features as input and returning predicted target quality-variables.

\item
\emph{Define a decision rule converting target quality-variables into a recommendation.}
We denote it by $\phi$.

%%%According to the previously introduced notation, in our problem this rule has the form $\phi \colon \mathbb{R}^{n_q} \rightarrow R \times M$.

\item
Construct recommendation system using the fitted model and the decision rule.
We denote it by $a(S) = \phi(p(f(S)))$, where $a$ is a recommendation system.
\end{enumerate}

%%SHORT: application of such a rec.syst.
%! double ":"
Application of such system is straightforward: for a new task with a dataset $S$, calculate meta-features $f(S)$, predict target quality-variables $p(f(S))$, apply the decision rule to get a recommendation: $a(S) = \phi(p(f(S)))$.

\subsection{Recommendation System Accuracy}
\label{sec:rec-accuracy}

Quality metric value of recommended method $a(S)$ on dataset $S$ is a natural estimate of recommendation system's accuracy on this dataset.
However, there are some important aspects to be considered.
%! "aspect" is not a perfect word here
%! rewrite sentence

Firstly, higher values of this metric do not imply better recommendation. It is possible that all the considered methods 
perform well on this particular dataset and $a(S)$ provide the lowest quality among them.
Therefore quality values of all the considered methods should be taken into account to estimate accuracy of recommendation.

One possible way to implement this idea was suggested by \citet{rec-clf-using-knn}.
They introduce the Recommendation Accuracy ($R\!A$) metric:
$$
R\!A(S) = \frac{Q_{a(S)} - Q_{\txt{worst}(S)}}{Q_{\txt{best}(S)} - Q_{\txt{worst}(S)}}.
$$
Here $Q_{a(S)}$ is a quality metric value of recommended method $a(S)$ on dataset $S$,
$Q_{\txt{best}(S)}$ and $Q_{\txt{worst}(S)}$ are the highest and the lowest quality metric values among all considered methods on dataset $S$.
Apparantly, $R\!A(S) \in [0, 1]$, and the higher value it has, the better the recommendation for dataset~$S$ is.
Average value of $R\!A(S)$ on all datasets $S \in B$ can be used to estimate overall accuracy of the recommendation system.

Secondly, the estimate of recommendation system's efficiency should be based not only on accuracy of recommended methods, but also on the runtime.
Indeed, if the runtime is ignored, the exhaustive search would be regarded as the best recommendation system.
However, as already mentioned, the exhaustive search can take a lot of time, so it does not fulfill the initial purpose of recommendation system.

%!refs to works
In most of previous works, researchers took runtime of recommendation systems into account by setting restrictions on their computatitonal complexity.
That is, they considered only computationally cheap meta-features and meta-learning models,  
which automatically results in low runtime of recommendation systems.

%! The other way: include runtime into metrics

\subsection{Related Work on Recommendation Systems}
\label{sec:rec-examples}

Below we give an overview of recommendation systems from previous papers by \citet{rec-fss}, \citet{rec-active}, \citet{rec-clf-ranking-using-knn}, \citet{rec-clf-using-knn}, \citet{rec-text-clf}, \citet{rec-clf-using-rules}, \citet{rec-clustering}, \citet{rec-ts}.
All of them fit into the template described in subsection~\ref{sec:rec-template}, so we go through its steps and show how they are implemented in each work.

%%%For example, in this paper we aim to build system recommending resampling method and multiplier, recommendation system described 
The recommendation system by \citet{rec-fss} recommends a ranking of feature selection methods.
The systems by \citet{rec-active, rec-clf-ranking-using-knn} recommend a ranking of classification models.
The systems by \citet{rec-clf-using-knn, rec-text-clf, rec-clf-using-rules} recommend a single classification model.
The system built by \citet{rec-clustering} recommends a ranking of clustering algorithms.
The system by \citet{rec-ts} recommends single time-series prediction model.

Researchers usually include in meta-features (see \citet{rec-clf-using-knn, rec-text-clf, rec-clustering, rec-ts}), general information about the task and the training dataset, metric, statistical, information-theoretical properties of the dataset, \emph{landmarks} (fast estimates of quality, see \citet{rec-active, rec-ml-concepts}).
%%%Note that meta-features shouldn't be computationally expensive because it would strongly increase time of recommendation providing (see section~\ref{sec:rec-apply-gen}), which is critical for recommendation systems.

The exact form of quality-variables and the meta-learning problem was defined in various ways.
For example,
\citet{rec-text-clf, rec-clf-ranking-using-knn} formulated a regression problem of the classification score prediction.
\citet{rec-fss} formed target variable from a score of a feature selection algorithm. \citet{rec-clf-using-rules} formulated a multi-class classification problem with labels coding the number of the best method.

Apparently, the formulated meta-learning problems could be solved by different means.
The most widespread were $k$-nearest-neighbors-based methods considered by \citet{rec-fss, rec-active, rec-clf-using-knn, rec-clf-ranking-using-knn}.
\citet{rec-text-clf} considered linear regression as one of the models, \citet{rec-clf-using-rules} used rule-based learning algorithm C$5.0$, \citet{rec-clustering} used SVM to solve the classification problem.

\section{Resampling Recommendation System Construction}
\label{sec:res-recsyst-construct}

%rewrite
In this section, we describe our implementation of the recommendation system template for the resampling selection problem.
We propose two approaches to construct resampling recommendation systems.
Both of them consider the same set of meta-features and have the same forms of recommendations,
so interfaces of resulting recommendation systems are identical.
The approaches differ in meta-learning problem formulation,
which results in different inner structure of recommendation systems.
Therefore we organize the description as follows.
In the first few subsections we describe things common for both approaches:
form of recommendation, meta-features, some quality-variables.
Then, starting from the meta-learning problem formulation, we describe the approaches separately.

\subsection{Purpose and Form of Recommendations}
\label{sec:resrec-purpose}
We consider binary classification tasks on imbalanced datasets.
According to section~\ref{sec:imb-notation}, to solve this task, one has to select resampling method $r$, multiplier $m$ and classification model $h$.
Let $h$ be fixed.
Let sets of considered resampling methods $R=\{r^0,\ r^1, \ldots, r^{n_1}\}$ (including no-resampling $r^0$) and multipliers $M=\{m_1, \ldots, m_{n_2}\}$ be also fixed.
We are going to construct a recommendation system $a$. It will take as input a dataset $S$ 
and recommend method $r \in R$ and multiplier $m \in M$ 
for resampling of this dataset before applying classification model $h$ on it ($a(S) = (r, \ m)$).

Note that $h$, $M$ and $R$ are not specified yet since the details of recommendation system algorithm do not depend on it (however, they will be specified in the experimental section~\ref{sec:exp-res-recsyst}).
In other words, we describe an approach 
which is scalable to different sets of considered classification models, resampling methods and multipliers.

\subsection{Classification Quality Metric}
\label{sec:clf-quality-metric}

Let some classification quality metric $Q$ (see section~\ref{sec:imb-notation}) be fixed.
We consider $k$-fold cross-validation estimate $Q^{kCV}_\txt{av}(h, r_m, S)$ as the quality measure of resampling method $r$ with multiplier $m$ applied to dataset $S$.
The process of recommendation system construction is independent of choice of $Q$ and $k$, so they are specified later in section~\ref{sec:exp-res-recsyst}.

\subsection{Recommendation Accuracy Metric}
\label{sec:res-rec-accuracy}

The value of $Q^{kCV}_\txt{av}(h, r_m, S)$ is also used to estimate the quality of a recommendation given by system $a$ for dataset $S$.
According to section~\ref{sec:rec-accuracy}, the time of providing a recommendation should be considered in recommendation system quality metric.
We achieve this by setting a restriction on the computational complexity.
That is, we consider only recommendation systems which require much less computational time than the exhaustive search
and we aim to construct one which would give higher value of $Q^{kCV}_\txt{av}(h, r_m, S)$ for any given dataset $S$.

\subsection{Meta-features}
\label{sec:res-recsyst-metafs}
We use the following list of meta-features: number of objects, number of features, objects-features ratio ($|S| / d$), reversed imbalanced ratio ($1 / I\!R$, see section~\ref{sec:imb-notation}), distance between centers of classes, minimal and maximal absolute eigenvalue of covariance matrix for each of two classes separately, minimal and maximal value of skewness among all features for each class separately, minimal and maximal $p$-value of skewness normality test among all features for each class separately, minimal and maximal value of kurtosis among all features for each class separately, minimal and maximal $p$-value of kurtosis normality test among all features for each class separately. We also use values of these meta-features in the logarithmic scale.

\subsection{Quality-variables}
\label{sec:res-recsyst-qvars}

%During learning of recommendation system,
For each dataset $S$, resampling method $r \in R$ (including no-resampling), resampling multiplier $m \in M$ we perform $k$-fold cross-validation and obtain $k$-dimensional vector of quality metric values $Q^{kCV}(h, r_m, S)$ (one value for each cross-validation iteration).
Below we describe how its components are organized into quality-variables for each dataset $S$.

The first group of quality-variables is connected with the performance of classification without resampling.
There is only one such variable: $q_0^{\txt{mean}}(S)$, which is an average of components of $Q^{kCV}(h, r^0, S)$.
%\item
%$q_0^{\txt{std}}(S)$, which is a standard deviation of components of $Q^{kCV}(h, r^0, S)$.
The second  group of quality-variables describe the classification performance with all considered ways of resampling.
That is, for each $r \in \{r^1, \ldots, r^{n_1}\}$ and for each $m \in M$ we compute:
\begin{itemize}
\item
$q_{r, m}^{\txt{mean}}(S)$, an average of components of $Q^{kCV}(h, r_m, S)$.
%\item
%$q_{r, m}^{\txt{std}}(S)$, which is a standard deviation of components of $Q^{kCV}(h, r_m, S)$.
\item
%ref: T-test
$q_{r, m}^{\txt{pval}}(S)$, which is a $p$-value of the t-test with null hypothesis that mean of $Q^{kCV}(h, r_m, S)$ is not greater than mean of $Q^{kCV}(h, r^0, S)$.
Thus, the lower this $p$-value is, the better the score of resampling $r_m$ is (compared to no-resampling).
\item
$q_{r, m}^{\txt{pvalw}}(S) = \max\limits_{m' \colon |m' - m| < \epsilon} q_{r, m'}^{\txt{pval}}(S)$ for some fixed $\epsilon$.  This is the maximum of $p$-values in $\epsilon$-window around $m$.
\end{itemize}
In the third group, quality-variables describe the performance of each resampling method $r \in \{r^1, \ldots, r^{n_1}\}$ used with the multiplier which is the best in some sense.
%terms of some previously introduced quality-variable.
There are many ways to define which multiplier is the best,
but our recommendation systems are based on the  one with the lowest maximum of $p$-values in $\epsilon$-window around it: 
\[m_{r}^{*\txt{minpvalw}}(S) = \min\limits_{m \in M} q_{r, m}^{\txt{pvalw}}(S).\]
%\end{itemize}

Thus for each $r \in \{r^1, \ldots, r^{n_1}\}$ 
%and for each $w \in \{\txt{maxmean}, \txt{minpval}, \txt{minpvalw}\}$ 
we include:

\begin{itemize}
\item
$q_{r, \txt{minpvalw}}^{\txt{mean}}(S) = q_{r, m^*}^{\txt{mean}}(S)$,

%\item
%$q_{r, \txt{minpvalw}}^{\txt{std}}(S) = q_{r, m^*}^{\txt{std}}(S)$, 
%where $m^*= m_{r}^{*\txt{minpvalw}}(S)$.

\item
$q_{r, \txt{minpvalw}}^{\txt{pval}}(S) = q_{r, m^*}^{\txt{pval}}(S)$, 
where $m^*= m_{r}^{*\txt{minpvalw}}(S)$.
\end{itemize}

\subsection{Approach No. 1}
\label{sec:res-recsyst-one}

\noindent This approach requires two additional parameters: 
\begin{itemize}
\item
$\alpha \in (0,1)$, which is a significance level for the test that no-resampling is better than resampling;

\item
$\epsilon > 0$, which is a half-width of window for multiplier (see previous subsection).
\end{itemize}
Let $\alpha$ and $\epsilon$ be fixed.

In addition, for each $r \in \{r^1, \ldots, r^{n_1}\}$, $m \in M$ one more quality-variable is introduced:
\begin{equation*}
y_{r, m}(S) = 
\begin{cases}
   1, & \mbox{if } q_{r, m}^{\txt{pval}}(S) < \alpha,\\
   0, & \mbox{otherwise.}
 \end{cases}
\end{equation*}
%! Actually, pvalw should be used here
%! Thus this comment should be rewritten respectively
Apparently, it represents the result of the above-mentioned test: it equals to $1$ if and only if the null-hypothesis is rejected with significance level $\alpha$ (i.e., resampling $(r, m)$ is better than no-resampling on dataset $S$).

For each $r \in \{r^1, \ldots, r^{n_1}\}$, $m \in M$ the following meta-learning problem is formulated.
Each meta-example $S$ is assigned to the class $0$ or to the class $1$ according to the value of $y_{r, m}(S)$.
The problem is to predict class label of meta-example $S$ using only its meta-features $f(S)$ and provide a probability estimate for this prediction.
Informally speaking, the meta-learning problem is to predict whether dataset $S$ is worth being resampled with method $r$ and multiplier $m$ or not.

%%%use meta-features $f(S)$ of meta-example $S$ to determine probabilities of belonging $S$ to these two classes.

Solving these meta-learning problems, we obtain a classifier for each $r$ and $m$.
It outputs $\hat{y}_{r,m}(f(S))$, which denotes the estimated class label for dataset $S$,
and $\hat{p}_{r,m}(f(S))$, which denotes the estimated probability that $S$ actually belongs to class~$1$.
These estimates are used to form a recommendation in the following natural way:
\begin{equation*}
a(S) = 
\begin{cases}
   \argmax\limits_{r,m} \hat{p}_{r,m}(f(S)), & \mbox{if } \max\limits_{r,m} \hat{y}_{r,m}(f(S)) = 1,\\
   (r^0, 1.0), & \mbox{otherwise.}
\end{cases}
\end{equation*}
Thus, if there is some kind of resampling which is significantly better than no-resampling for dataset $S$ than recommend the resampling with the highest probability of being better; otherwise, recommend no-resampling.

\subsection{Approach No. 2}
\label{sec:res-recsyst-two}

\noindent This approach requires two additional parameters: 
\begin{itemize}
\item
$\alpha \in (0,1)$, which is a significance level for the test that no-resampling is better than resampling;

\item
$\epsilon > 0$, which is a half-width of window for multiplier (see subsection \ref{sec:res-recsyst-qvars}).
\end{itemize}
Let $\alpha$ and $\epsilon$ be fixed. For each $r \in \{r^1, \ldots, r^{n_1}\}$, two additional quality-variables are introduced:
\begin{equation*}
y_{r}(S) = 
\begin{cases}
   1, & \mbox{if } \min\limits_{m} q_{r, m}^{\txt{pval}}(S) < \alpha,\\
   0, & \mbox{otherwise};
 \end{cases}\qquad
z_{r}(S) = \argmin\limits_m q_{r, m}^{\txt{pval}}(S).
\end{equation*}

Thus, $y_{r}(S)$ indicates whether resampling method $r$ can give a statistically significant quality improvement on dataset $S$ if the multiplier is chosen properly.
Variable $z_{r}(S)$ represents the best choice of multiplier for dataset $S$ resampled with method $r$.

For each resampling method $r \in \{r^1, \ldots, r^{n_1}\}$ we formulate two meta-learning problems. 

The first problem is to predict $y_{r}(S)$ for dataset $S$ using its meta-features $f(S)$.
This is a binary classification task, and it is very similar to the one stated in the previous subsection.
Informally, the problem is to determine whether dataset $S$ is worth being resampled using method $r$ granted that multiplier is chosen properly.

The second problem is to predict $z_{r}(S)$ for dataset $S$ using its meta-features $f(S)$.
This is a regression task of predicting most appropriate multiplier $m$ for the resampling method $r$ applied to dataset $S$.

Having these problems solved, we obtain a classifier and a regression model for each $r \in \{r^1, \ldots, r^{n_1}\}$.
We denote predicted $y_r$ as $\hat{y}_r(S)$, probability estimate for this prediction as $\hat{p}_r(S)$ and predicted $z_r$ as $\hat{z}_r(S)$.
These estimates are used to form a recommendation in the following way: $a(S) = (r^*, m^*)$, where
\begin{equation*}
r^*(S) = 
\begin{cases}
   \argmax\limits_{r} \hat{p}_{r}(f(S)), & \mbox{if }\max\limits_{r} \hat{y}_{r}(f(S)) = 1,\\
   r^0, & \mbox{otherwise;}
\end{cases}\qquad
m^*(S) = 
\begin{cases}
   z_{r^*}(S), & \mbox{if } \max\limits_{r} \hat{y}_{r}(f(S)) = 1,\\
   1.0, & \mbox{otherwise;}
\end{cases}
\end{equation*}

Informally, if there is some resampling method (with properly chosen multiplier) which can be significantly better than no-resampling, then we recommend this method and the most appropriate multiplier for it.
Otherwise, we recommend no-resampling.

\section{Resampling Recommendation System Learning and Evaluation}
\label{sec:exp-res-recsyst}

In this section, we describe the experimental part of our research: testing the performance of the classification models with resampling on various imbalanced datasets, forming the meta-examples set, learning resampling recommendation systems and assesing their quality.\footnote{Full source code and the datasets (used in the experiment) are available online in Public GitHub repository \url{https://github.com/papart/res-recsyst}.}

\subsection{Preparation of Meta-examples Set}
\label{sec:exp-meta-examples-prep}

In order to learn and evaluate resampling recommendation systems described in section~\ref{sec:res-recsyst-construct}, we prepared a set of meta-examples according to the general recommendation system template (see section~\ref{sec:rec-template}).
Although we have already specified some aspects of meta-examples construction (see, for example, sections \ref{sec:res-recsyst-metafs} and \ref{sec:res-recsyst-qvars}), some other aspects and technical details still remain undefined.
In this subsection, we provide a thorough description of meta-examples set preparation.

\subsubsection{Datasets}
\label{sec:exp-data}
We used two pools of datasets:
artificial ($\sim\! 1000$ datasets) and real with ($\sim\! 100$ datasets).\footnote{Real datasets are availaible online at \url{http://sci2s.ugr.es/keel/imbalanced.php} and \url{http://homepage.tudelft.nl/n9d04/occ/index.html}}.
Artificial datasets were drawn from a Gaussian mixture distribution.
Each of two classes is a Gaussian mixture with not more than $3$ components.
Number of features varies from $6$ to $40$, sizes of datasets varies from $200$ to $1000$, $I\!R$ from $0.05$ to $0.35$.
For more details on artificial data generation check the repository \url{https://github.com/papart/res-recsyst}.
Real-world datasets came from different areas: biology, medicine, engineering, sociology.
All features are numeric or binary, their number varies from $3$ to $1000$.
Dataset size varies from $200$ to $1000$, and $I\!R$ varies from $0.02$ to $0.75$.

\subsubsection{Classification Models}
\label{sec:exp-clf}

We use Decision trees, $k$-Nearest neighbors ($k=5$), and Logistic regression with $\ell_1$ regularization as a classification model ($h$). Optimal parameters of classification models were selected by cross-validation.

\subsubsection{Resampling Methods and Multipliers}
\label{sec:exp-res-methods-mult}

Set $R$ of resampling methods consists of no-resampling, Bootstrap, RUS and SMOTE with $k \in \{1,3,5,7\}$.
Values of resampling multiplier $m$ are from $1.25$ to $10.0$ with step~$=0.25$.

\subsubsection{Classification Quality Evaluation}
\label{sec:exp-clf-qmetric}
%!reference
The area under precision-recall curve is used as a classification quality metric $Q$.
We estimate the quality of classification with resampling as it was described in section~\ref{sec:imb-notation} (with number of cross-validation folds $k=20$).
Thus, for each dataset $S$, resampling method $r \in R$, resampling multiplier $m \in M$ and classification model $h$ we obtain $k=20$-dimensional vector of quality metric values $Q^{kCV}(h, r_m, S)$.

\subsubsection{Meta-features}
\label{sec:exp-clf-metafs}

For each dataset $S$, we calculate meta-features listed in section~\ref{sec:res-recsyst-metafs}.

\subsubsection{Quality-variables}
\label{sec:exp-clf-qvars}

For each dataset $S$ we use quality metric values $Q^{kCV}(h, r_m, S)$ to calculate quality-variables (see section~\ref{sec:res-recsyst-qvars}).
We set half-width of multiplier window $\epsilon$ to $0.75$.

% Should I write something about it?
%\subsection{Meta-Feature Selection}

% Should I write something about it?
% data preprocessing --- a better name for the subsection
%\subsection{Meta-Feature Scaling}

% Should I write something about it?
%\subsection{Outliers among Meta-examples}

\subsection{Learning and Quality Evaluation Process}

According to \citet{res-influence}, the efficiency of a resampling method on a particular task depends on the classification model.
Therefore we treat resampling recommendation problem for each classification model $h$~(see~\ref{sec:exp-clf}) separately.
%! 'approach' is not the best word here

For each $h$ we construct two recommendation systems according to the approaches described in subsections~\ref{sec:res-recsyst-one} and \ref{sec:res-recsyst-two}.
The learning process has been already described in these subsections (except for values of parameters, selected meta-features and models for meta-problem solving; they will be specified in the following subsection~\ref{sec:exp-res-recsyst-models}).
However, apart from learning recommendation systems, we need to evaluate their quality, which has not been discussed yet.

In order to evaluate quality of resampling recommendation systems on all availiable datasets, we use $k'$-fold cross-validation ($k'=10$).
That is, we randomly split the bank of datasets $B$ into $k'$ subsets $B_1, \ldots, B_{k'}$ of roughly equal size.
Then, for each $j=\overline{1,k'}$ we perform two steps.
\begin{enumerate}
\item
\textit{Training.} 
We learn a recommendation system $a_j$ as described in section~\ref{sec:res-recsyst-one} on all datasets except those from $B_j$.
Preparation of meta-examples from datasets is described in details in section~\ref{sec:exp-meta-examples-prep}.
Selected meta-features and models for recommendation system construction are provided in section~\ref{sec:exp-res-recsyst-models}.

\item
\textit{Testing. }
We evaluate recommendation quality $Q_{a_j(S)}(S)$ values on all datasets $S \in B_j$.
\end{enumerate}

As a result, for each dataset $S \in B$ we obtain the quality of recommendation achieved by the system.
% (?)Actually, we evaluate the approach to build the recommendation system, not the recsyst itself.
These quality values are used to evaluate overall performance of recommendation systems, see section~\ref{sec:exp-res-recsyst-accuracy} for details.

%! Add: Since statistical properties differ ...
% We run separate cross-validation for real and artificial datasets.

\subsection{Meta-features and Models}
\label{sec:exp-res-recsyst-models}

In this subsection we specify the detais of the considered methods and approaches.

\subsubsection{Rec. System No. 1 for Decision Tree Classifier}

\begin{itemize}
\item
Classification model for solution of the meta-learning problem: AdaBoost with Decision Tree as a base classifier and the number of estimators set to 10.

\item
Meta-features: reversed imbalanced ratio, distance between class centers, number of objects, minimal absolute eigenvalue of covariance matrix of the major class, maximal $p$-value of kurtosis normality test among all features in the minor class.
%\item
Significance level: $\alpha=0.05$.
\end{itemize}

\subsubsection{Rec. System No. 2 for Decision Tree Classifier}

\begin{itemize}
\item
Classification model for solution of the meta-learning problem: AdaBoost with Decision Tree as a base classifier and the number of estimators set to 10.

\item
Regression model for solution of the meta-learning problem: AdaBoost with Decision Tree Regressor as a base model and the number of estimators set to 10.

\item
Meta-features: reversed imbalanced ratio, distance between class centers.
%\item
Significance level: $\alpha=0.05$.
\end{itemize}

\subsubsection{Rec. System No. 1 for $k$-Nearest Neighbors}
\begin{itemize}
\item Classification model for solution of the meta-learning problem: AdaBoost with Decision Tree as a base classifier and the number of estimators set to 10.

\item
Meta-features: reversed imbalanced ratio, distance between class centers, number of objects, minimal absolute eigenvalue of covariance matrix of the major class, maximal $p$-value of kurtosis normality test among all features in the minor class.
%\item
Significance level: $\alpha=0.05$.
\end{itemize}

\subsubsection{Rec. System No. 2 for $k$-Nearest Neighbors}
\begin{itemize}\item Classification model for solution of the meta-learning problem: AdaBoost with Decision Tree as a base classifier and the number of estimators set to 10.

\item
Regression model for solution of the meta-learning problem: AdaBoost with Decision Tree Regressor as a base model and the number of estimators set to 10.

\item
Meta-features: reversed imbalanced ratio, distance between class centers.
%\item
Significance level: $\alpha=0.05$.
\end{itemize}

\subsubsection{Rec. System No. 1 for $\ell_1$ Logistic Regression}
\begin{itemize}
\item
Classification model for solution of the meta-learning problem: Logistic Regression with $\ell_1$ regularization.

\item
Meta-features: reversed imbalanced ratio, distance between class centers, number of objects, minimal absolute eigenvalue of covariance matrix of the major class, minimal and maximal $p$-value of kurtosis normality test among all features in the minor class, minimal and maximal $p$-value of skewness normality test among all features in the minor class.
%\item
Significance level: $\alpha=0.3$.
\end{itemize}

\subsubsection{Rec. System No. 2 for $\ell_1$ Logistic Regression}

\begin{itemize}
\item
Classification model for solution of the meta-learning problem: AdaBoost with Decision Tree as a base classifier and the number of estimators set to 10.

\item
Regression model for solution of the meta-learning problem: AdaBoost with Decision Tree Regressor as a base model and the number of estimators set to 10.

\item
Meta-features: reversed imbalanced ratio, distance between class centers.
%\item
Significance level: $\alpha=0.05$.
\end{itemize}

\subsection{Resampling Recommendation System Quality Assessment}
\label{sec:exp-res-recsyst-accuracy}

\subsubsection{Accuracy Metric}
\label{sec:exp-res-recsyst-metric-graphs}
We use Recommendation Accuracy metric (see section~\ref{sec:rec-accuracy}) to estimate accuracy of resampling recommendation system $a$ on dataset $S$:
\[
R\!A_a(S) = \frac{ q_{a(S)}^{\txt{mean}}(S) - \min\limits_{r, m} q_{r,m}^{\txt{mean}}(S)}{\max\limits_{r, m} q_{r,m}^{\txt{mean}}(S) - \min\limits_{r, m} q_{r,m}^{\txt{mean}}(S)}.
\]

We plot the  empirical distribution function of $R\!A$ metric to represent its values on all datasets in a convenient way.
That is, for each value of the metric $x = R\!A_a(S), S \in B$ we calculate share of datasets $y(x) = \frac{1}{|B|}  \left| \left\{S \in B \colon R\!A_a(S) < x \right\}\right|$ and plot $y(x)$ versus $x$.
The distribution function is non-decreasing and its graph is contained within a unit square.
The best possible recommendation system gives $R\!A(S)=1$ on each dataset $S$, so its distribution function starts in $(0,0)$, then goes to the point $(1,0)$ and moves upward to the point $(1,1)$.
Therefore the closer to the bottom-right corner $(1,0)$ the distribution function graph is located, the better the recommendation system is.

We also provide average value of $R\!A$ metric on all datasets in order to give a value describing overall accuracy of recommendation system $a$, that is, \[A R\!A_a = \frac{1}{|B|} \sum\limits_{S \in B} R\!A_a(S).\]

\subsubsection{Methods to Compare with}

It is reasonable to compare resampling recommendation systems with static strategies of resampling selection.
%why?
We consider the simplest ones: no-resampling; resampling using Bootstrap, RUS or SMOTE (with $k=5$) which provides balanced classes ($I\!R=1$) in the resulting dataset.

Each of these strategies (as well as each resampling method with any fixed multiplier) can be regarded as a trivial recommendation system.
Therefore it is possible to apply recommendation system evaluation methodology from section~\ref{sec:exp-res-recsyst-metric-graphs} directly to them.
This allows us to compare resampling recommendation system we built with these strategies in the same terms.
More specifically, we are going to compare $R\!A$ empirical distribution functions and values of $A R\!A$.

\subsection{Results}

\begin{figure}[!t]

\begin{minipage}{0.48\textwidth}
\includegraphics[scale=0.31]{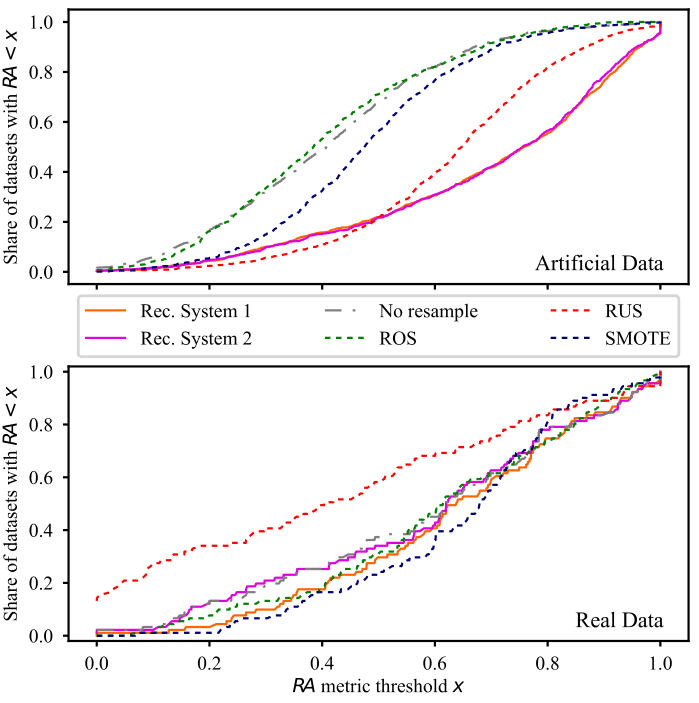}
\end{minipage}\hfill
\begin{minipage}{0.48\textwidth}
\begin{tabular}{lcc}
\toprule
{} &  \shortstack{Mean $R\!A$ value\\(Artificial Data)} &  \shortstack{Mean $R\!A$ value\\(Real Data)} \\
\midrule
Rec. System 1 &          0.6942 &    0.6381 \\
Rec. System 2 &          0.6900 &    0.5972 \\
No resample   &          0.4081 &    0.5878 \\
ROS, EqS      &          0.4024 &    0.6084 \\
RUS, EqS      &          0.6326 &    0.4216 \\
SMOTE, EqS    &          0.4791 &    0.6407 \\
\bottomrule
\end{tabular}

%\vspace{2mm}

\caption{Recommendation systems vs static strategies. Empirical distribution functions of $R\!A$ and average values of $R\!A$.
Classifier: Decision Tree. 
Metric: area under the Precision-Recall curve.}
\label{fig: ra-cdf-mean-dtree}
\end{minipage}
\end{figure}

\begin{figure}[!htb]
\begin{minipage}{0.48\textwidth}
\includegraphics[scale=0.31]{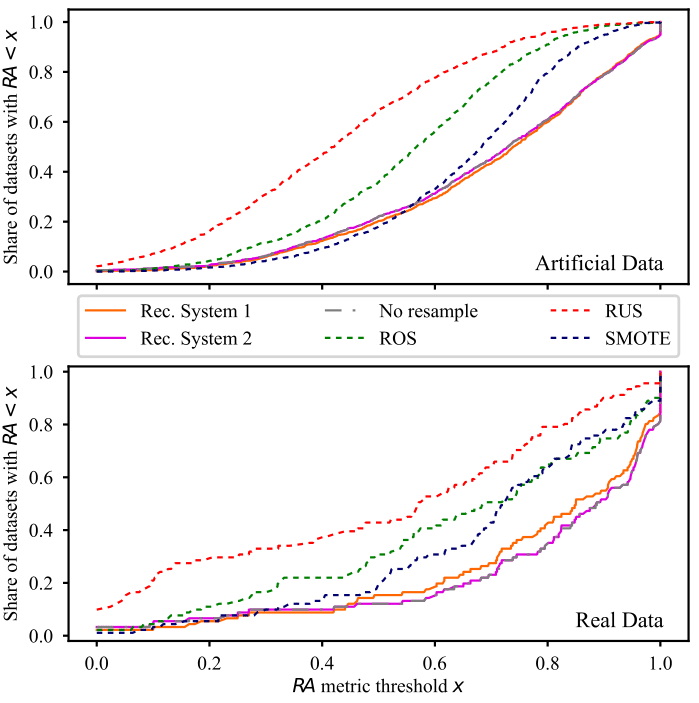}

%\vspace{2mm}
\begin{tabular}{lcc}
\toprule
{} &  \shortstack{Mean $R\!A$ value\\(Artificial Data)} &  \shortstack{Mean $R\!A$ value\\(Real Data)} \\
\midrule
Rec. System 1 &          0.6986 &    0.7830 \\
Rec. System 2 &          0.6907 &    0.7925 \\
No resample   &          0.6910 &    0.7915 \\
ROS, EqS      &          0.5527 &    0.6431 \\
RUS, EqS      &          0.4228 &    0.5033 \\
SMOTE, EqS    &          0.6543 &    0.6883 \\
\bottomrule
\end{tabular}

\vspace{2mm}

\caption{Recommendation systems vs static strategies. Empirical distribution functions of $R\!A$ and average values of $R\!A$.
Classifier: $k$ Nearest Neighbors. 
Metric: area under the Precision-Recall curve.}
\label{fig: ra-cdf-mean-knn}
\end{minipage}\hfill
\begin{minipage}{0.48\textwidth}

\includegraphics[scale=0.31]{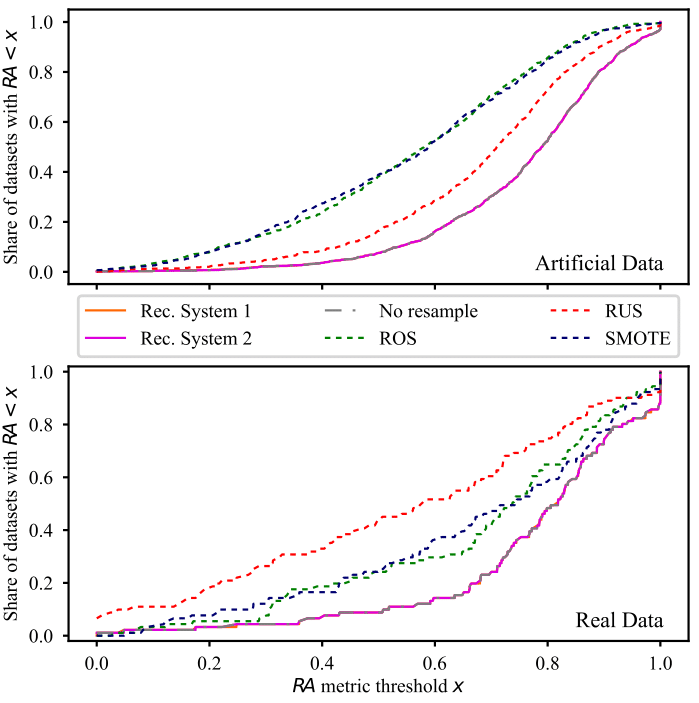}

%\vspace{2mm}

\begin{tabular}{lcc}
\toprule
{} &  \shortstack{Mean $R\!A$ value\\(Artificial Data)} &  \shortstack{Mean $R\!A$ value\\(Real Data)} \\
\midrule
Rec. System 1 &          0.7571 &    0.7724 \\
Rec. System 2 &          0.7573 &    0.7724 \\
No resample   &          0.7573 &    0.7724 \\
ROS, EqS      &          0.5558 &    0.6758 \\
RUS, EqS      &          0.6785 &    0.5435 \\
SMOTE, EqS    &          0.5554 &    0.6697 \\
\bottomrule
\end{tabular}

\vspace{2mm}

\caption{Recommendation systems vs static strategies. Empirical distribution functions of $R\!A$ and average values of $R\!A$.
Classifier: Logistic Regression with $\ell_1$ regularization. 
Metric: area under the Precision-Recall curve.}
\label{fig: ra-cdf-mean-logreg}
\end{minipage}
\end{figure}

%The following figures demonstrate performance of resampling recommendation systems as described in subsection~\ref{sec:exp-res-recsyst-accuracy}.
Performance of systems recommending resampling for usage with Decision tree classification model is shown in Figure~\ref{fig: ra-cdf-mean-dtree}.
Here both recommendation systems outperform other resampling selection strategies on artificial datasets.
However, they do not show a distinct quality improvement on real datasets:
SMOTE has a slightly higher average $R\!A$, 
ROS and no-resample show virtually the same performance as the recommendation systems.
This significant difference of performance on artificial and real data can be explained by the fact that the pool of artificial datasets is larger and less diverse than the pool of real ones.
We suppose that using a richer real data pool 
would lead to a more accurate meta-learning problem solution
and, consequently, higher performance of recommendation systems.

Figures~\ref{fig: ra-cdf-mean-knn} and \ref{fig: ra-cdf-mean-logreg} show performance of recommendation systems for $k$-Nearest neighbors classifier and $\ell_1$ Logistic regression, respectively.
In these cases, both recommendation systems achieve a significant quality improvement compared to static strategies ROS, RUS and SMOTE on artificial and real data.
However, they have essentially the same quality as no-resampling.
It turned out that the systems recommend no-resampling for most of datasets because their meta-features did not indicate that any resampling could be beneficial.
This result can be considered as a positive one since no-resampling has high $AR\!A$ value and outperforms other static strategies of resampling selection.
On the other hand, recommendation systems failed to recognize those few classification tasks which would have been solved better with resampling.
We suppose that it is possible to improve detection of such datasets by introducing new meta-features and more accurately fitting of the meta-learning model.

\section{Conclusion}
The meta-learning approach has been successfully applied to various method selection tasks in previous works.
We managed to use this approach for recommendation of resampling in imbalanced classification tasks, which has not been done previously.
We proposed two natural ways to construct resampling recommendation system and implemented them.
The systems we built showed good results: in all cases they achieve $AR\!A > 0.6$, which is better than random choice of resampling; for some cases they outperform static strategies of resampling selection.

Besides, we can note several directions of further research which can lead to improvements of recommendation systems quality.
First, one can try to introduce new meta-features which are capable of taking into account the specific nature of imbalanced classification tasks.
Second, meta-features can be selected more carefully.
Third, it is possible to find more accurate models for meta-learning problem solving.
Finally, one can use a bigger and more diverse dataset to learn a recommendation system.

\end{document}